\documentclass[runningheads]{llncs}
\usepackage{graphicx}

\usepackage{tikz}
\usepackage{comment}
\usepackage{amsmath,amssymb} 
\usepackage{color}

\usepackage{color,xcolor}
\usepackage{epsfig}
\usepackage{graphicx}
\usepackage{algorithm,algorithmic}

\usepackage{adjustbox}
\usepackage{array}
\usepackage{booktabs}
\usepackage{colortbl}
\usepackage{float,wrapfig}
\usepackage{framed}
\usepackage{hhline}
\usepackage{multirow}
\usepackage{tabularx}
\usepackage{subcaption} 
\usepackage[font=small]{caption}
\usepackage[percent]{overpic}

\usepackage{amsmath,amsfonts,amssymb}
\usepackage{bm}
\usepackage{nicefrac}
\usepackage{microtype}
\usepackage{contour}
\usepackage{courier}

\usepackage{changepage}
\usepackage{extramarks}
\usepackage{fancyhdr}
\usepackage{lastpage}
\usepackage{setspace}
\usepackage{soul}
\usepackage{xspace}
\usepackage{cuted}
\usepackage{fancybox}
\usepackage{afterpage}
\usepackage{enumerate}
\usepackage{paralist}
\usepackage{enumitem} 

\usepackage{url}
\usepackage{quoting}
\usepackage{epigraph}

\usepackage{comment}
\usepackage{pdfpages}

\makeatletter
\DeclareRobustCommand\onedot{\futurelet\@let@token\@onedot}
\def\@onedot{\ifx\@let@token.\else.\null\fi\xspace}

\makeatother

\newcommand{\X}{\mathcal{X}}
\newcommand{\F}{\mathcal{F}}
\newcommand{\T}{\mathcal{T}}

\newcommand{\LL}{\mathcal{L}}
\newcommand{\C}{\mathcal{C}}

\newcommand{\R}{\mathbb{R}}

\newcommand{\bb}{\boldsymbol{b}}
\newcommand{\bp}{\boldsymbol{p}}
\newcommand{\bs}{\boldsymbol{s}}

\newcommand{\bx}{\boldsymbol{x}}

\newcommand{\ff}{\boldsymbol{f}}
\newcommand{\bl}{\boldsymbol{l}}
\newcommand{\bc}{\boldsymbol{c}}
\newcommand{\bphi}{\boldsymbol{\phi}}
\newcommand{\btheta}{\boldsymbol{\theta}}

\definecolor{MyDarkBlue}{rgb}{0,0.08,1}
\definecolor{MyDarkGreen}{rgb}{0.02,0.6,0.02}
\definecolor{MyDarkRed}{rgb}{0.8,0.02,0.02}
\definecolor{MyDarkOrange}{rgb}{0.40,0.2,0.02}
\definecolor{MyPurple}{RGB}{111,0,255}
\definecolor{MyRed}{rgb}{1.0,0.0,0.0}
\definecolor{MyGold}{rgb}{0.75,0.6,0.12}
\definecolor{MyDarkgray}{rgb}{0.66, 0.66, 0.66}

\begin{document}
\pagestyle{headings}
\mainmatter
\def\ECCVSubNumber{100}  

\title{Multi-Frame to Single-Frame:\\Knowledge Distillation for 3D Object Detection} 

\titlerunning{Wang et al.}
%
\author{Yue Wang\inst{1} \and
Alireza Fathi\inst{2} \and
Jiajun Wu \inst{3} \and \\
Thomas Funkhouser\inst{2} \and
Justin Solomon\inst{1}}
\authorrunning{Wang et al.}
%
\institute{MIT \\
\email{\{yuewangx,jsolomon\}@mit.edu}
\and
Google \\
\email{\{alirezafathi,tfunkhouser\}@google.com}
\and
Stanford \\
\email{jiajunwu@cs.stanford.edu} 
}
\maketitle

\begin{abstract}
A common dilemma in 3D object detection for autonomous driving is that high-quality, dense point clouds are only available during training, but not testing. We use knowledge distillation to bridge the gap between a model trained on high-quality inputs at training time and another tested on low-quality inputs at inference time. In particular, we design a two-stage training pipeline for point cloud object detection. First, we train an object detection model on dense point clouds, which are generated from multiple frames using extra information only available at training time. Then, we train the model's identical counterpart on sparse single-frame point clouds with consistency regularization on features from both models. We show that this procedure improves performance on low-quality data during testing, without additional overhead. 
\end{abstract}
\section{Introduction}


Recent advances in 3D object detection are beginning to yield practical vision systems for autonomous driving. Given the realities of data collection and allowable inference time in this application, however, most relevant 3D object detection methods take static one-frame point clouds as input, which can be extremely sparse. The goal of matching object detection rates achievable from dense point clouds constructed by fusing multiple frames remains elusive. 
On the other hand, aggregating point clouds from multiple frames is inherently hard in dynamical settings. For example, Figure~\ref{fig:pointclouds} (b) shows a point cloud generated by combining 5 LiDAR scans, which exhibits significant motion blur; this blur---unavoidable in a moving environment---yields ambiguity in the object detection model. 

Point cloud object detection training data often comes in \emph{sequences}, with consistent labels that track moving objects between frames in each sequence; this extra information allows us to generate Figure~\ref{fig:pointclouds} (c), which shows an aggregated point cloud after accounting for the motion of individual objects. This tracking information, however, is generally unavailable at testing time. This mismatch between training and testing data leads to an unavoidable dilemma: It is desirable to build a model that gathers dense point clouds from multiple frames, but in the absence of fine-tuned registration procedures this data is only available at training time. 


In this paper, we introduce a two-stage training pipeline to bridge the gap between training on dense point clouds and testing on sparse point clouds. In principle, our method is an intuitive extension and application of knowledge distillation~\cite{Bucilua2006,knowledgedistillation} and learning using privileged information~\cite{VapnikV09,LopSchBotVap16,su2017adapting}. First, we train an object detection model on dense point clouds aggregated from multiple frames, using information available during training but not testing. Then, we train an identical model on single-frame sparse point clouds; we use the pre-trained model from the first stage to provide additional supervision for the second model. At test time, the second model makes predictions on single-frame point clouds, without extra inference overhead. Compared with a model trained solely on sparse point clouds, the proposed model benefits from distilled information gathered by the model trained on dense point clouds.

\section{Method}

\begin{figure}[t]  
  \centering
  
  \begin{tabular}{@{}ccc@{}}
 \includegraphics[width=0.3\columnwidth]{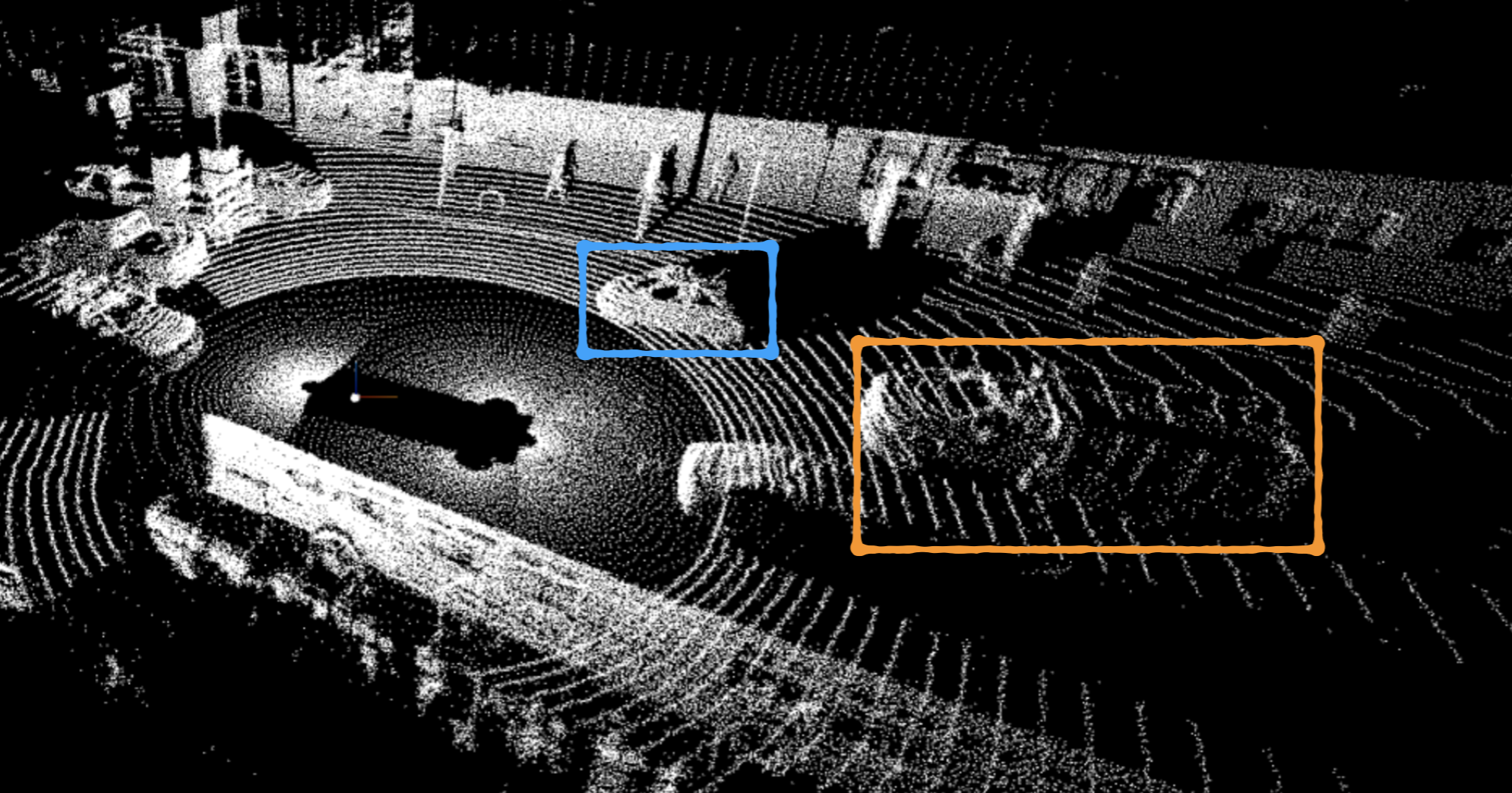} & 
    \includegraphics[width=0.3\columnwidth]{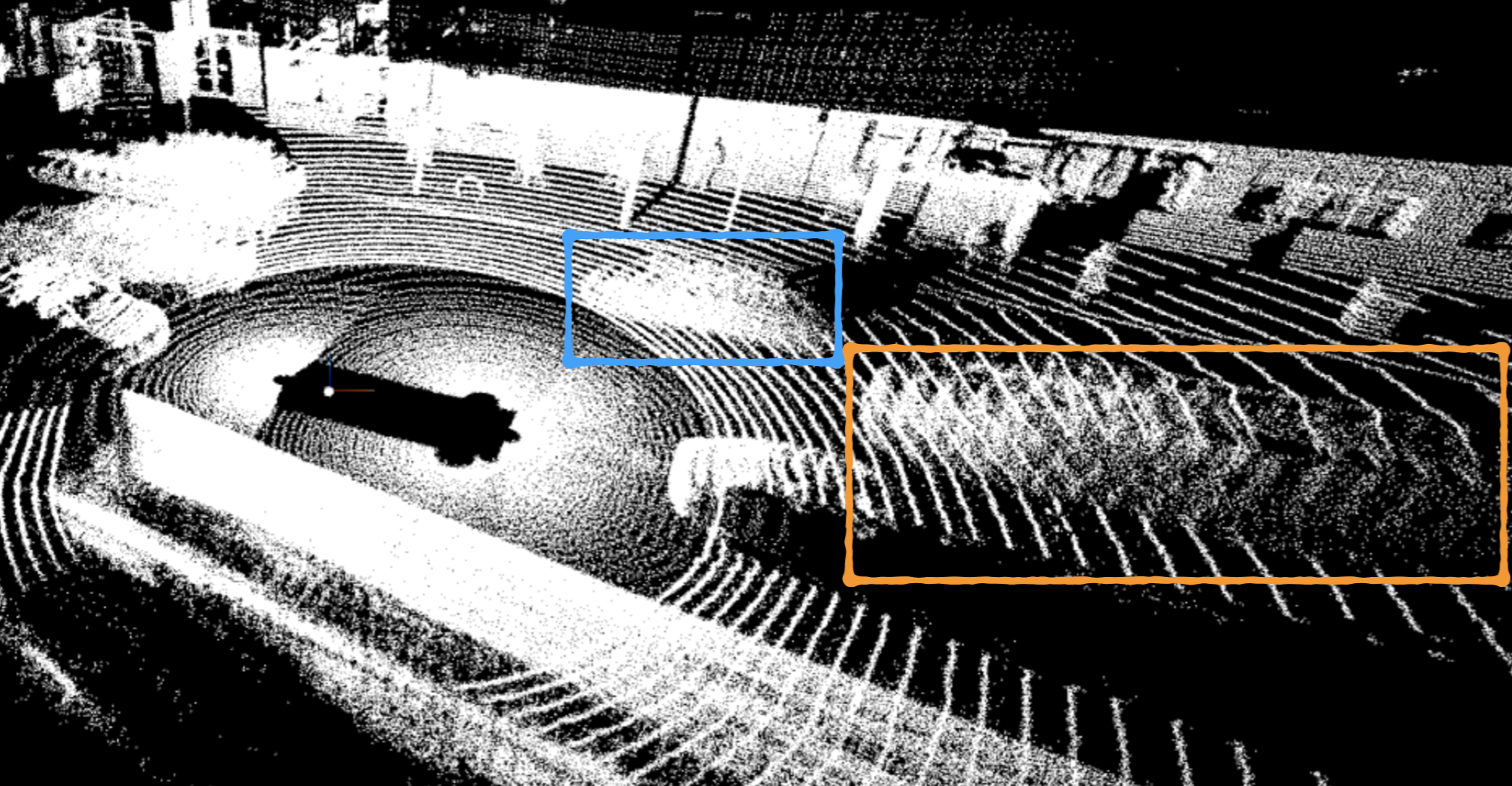} & 
    \includegraphics[width=0.3\columnwidth]{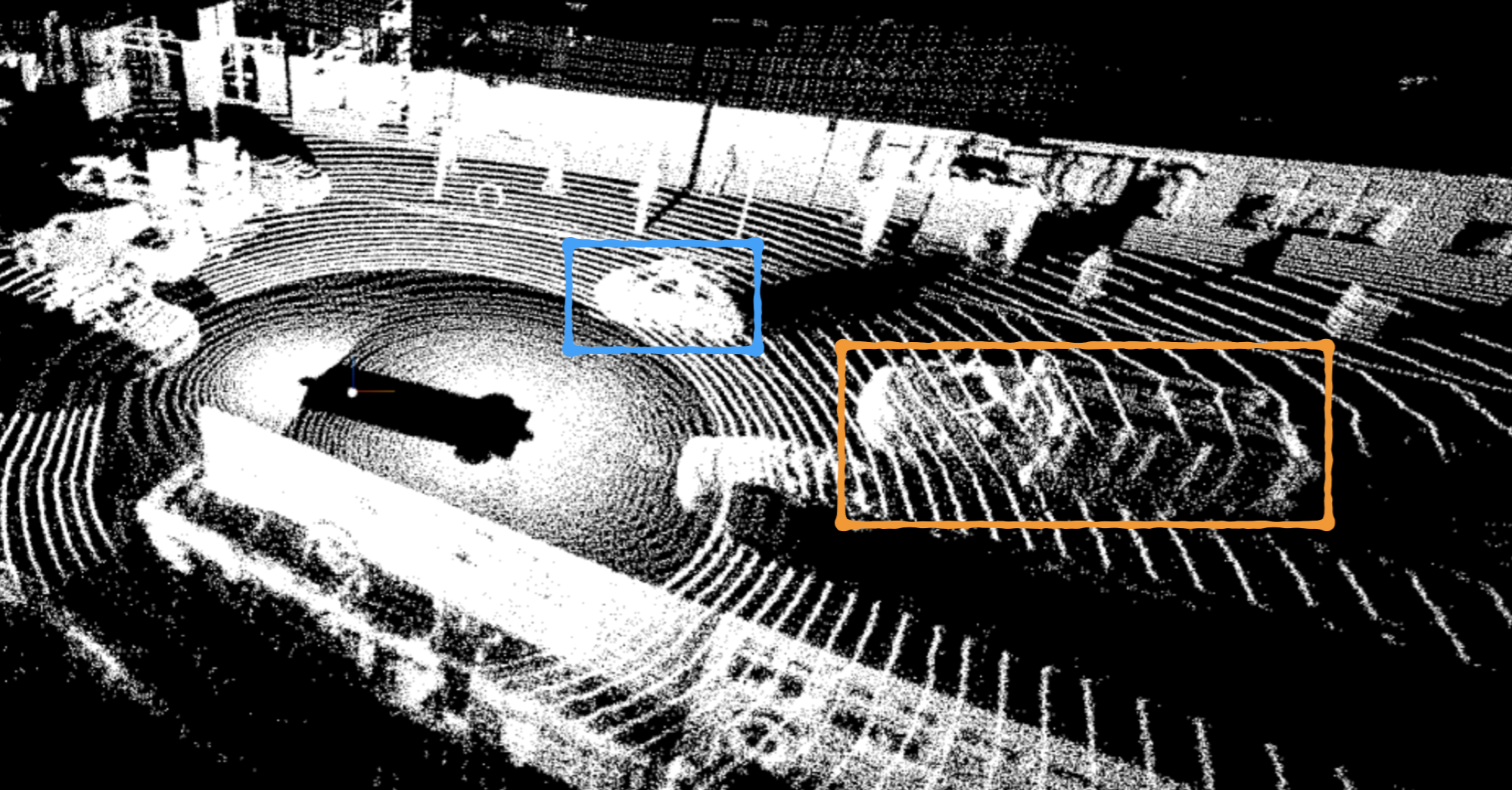}\\ 
(a) Single-frame & (b) Multi-frame w/o tracking &
(c) Multi-frame w/ tracking 
\end{tabular}
\vspace{-5pt}
  \caption{Visualizations of single/multi-frame point clouds. (a) single-frame point clouds; (b) multi-frame point clouds without bounding box tracking; (c) multi-frame point clouds with bounding box tracking. Two examples are labeled in blue and orange; the moving cars are extremely blurry in (b) while not in (c). \label{fig:pointclouds}  }
\end{figure}





We study how to use privileged information available at training time in the context of point cloud object detection. In contrast to existing works~\cite{su2017adapting,Lambert_2018_CVPR}, which use additional image information and require extra annotations, our model takes advantage of the privileged information that is naturally available in a sequence of point clouds. Our model bridges the gap between the single-frame model and the (impractical) multi-frame model, while maintaining efficiency. 

Consider a sequence of point clouds $\{\X_0, \X_1, \dots, \X_t, \dots, \X_T\}$ up to frame $T$ with per-frame pose 
$\{\T_0, \T_1, \dots, \T_t, \dots, \T_T\}$. Each frame is a point cloud $\X_t=\{\bx_t^0, \bx_t^1, \dots, \bx_t^i, \dots, \bx_t^N\} \subset \R^3$ 
with additional point-wise features $\F_t=\{\ff_t^0, \ff_t^1, \dots, \ff_t^i, \dots, \ff_t^N\} \subset \R^c$, such as reflectance and material.  
For simplicity of notation
, we assume the same number of points $N$ in each point cloud. 

In each frame $t$, a set of ground-truth bounding boxes $\{\bb_t^0, \bb_t^1, \dots, \bb_t^k, \dots, \bb_t^K\}$ is provided, where each bounding box is parameterized by its center location 
$\bp_t^k$, its size $\bs_t^k$, and its heading angle $\btheta_t^k$. For ease of presentation, assume $\{\bb_0^k, \bb_1^k, \dots, \bb_t^k \dots, \bb_T^k\}$ are bounding boxes of the $k$-th object over all frames; that is, we assume each frame in the sequence is labeled \emph{consistently}. We can take advantage of these bounding box annotations to generate dense point clouds as in Figure~\ref{fig:pointclouds} (c), but this is only possible at training time.

At test time, since we can no longer access the bounding box annotations, the inputs to the model are either single-frame point clouds (Figure~\ref{fig:pointclouds} (a)) or multi-frame point clouds (Figure~\ref{fig:pointclouds} (b)) without accounting for moving objects. So the model trained on the desired point clouds (Figure~\ref{fig:pointclouds} (c)) can never be deployed for inference.  To address this difference between training and testing, we propose a two-stage training pipeline. First, we produce dense point clouds by registering multiple frames and train an object detection model on dense point clouds. Then, we distill the previous model to a new model by leveraging both dense point clouds and sparse point clouds. Finally, we use the new model for testing on sparse point clouds without extra cost.

\section{Experiments}
First, we introduce the dataset, metrics, implementation details, and optimization details in~\S\ref{sec:detail}. Then, we show performance of the proposed method for vehicle and pedestrian detection on the Waymo Open Dataset~\cite{Sun2019ScalabilityIP} in~\S\ref{sec:result}. 

\subsection{Details}
\label{sec:detail}

\begin{figure}[t!]  
  \centering
    \includegraphics[width=\columnwidth]{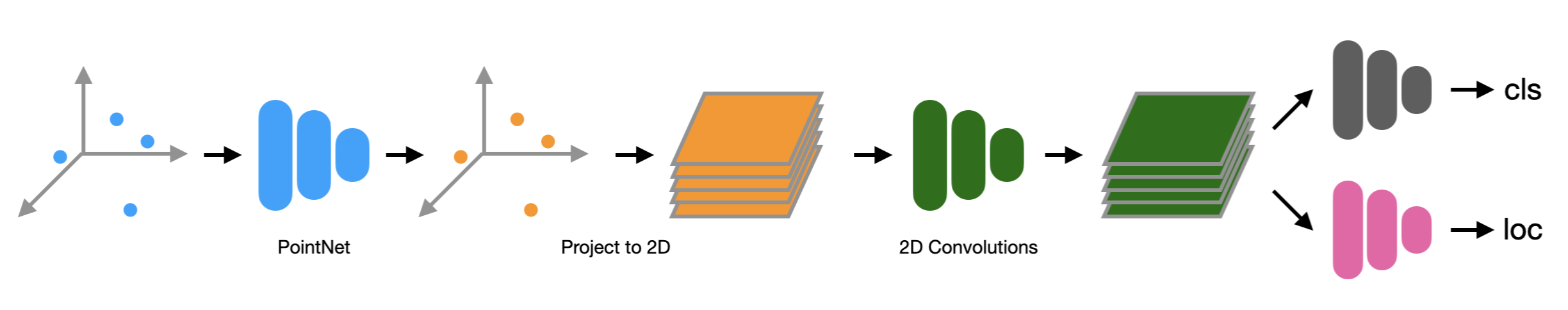}
   \vspace{-8pt}
  \caption{Our object detection model: first, point clouds are encoded using a lightweight PointNet; then, points are projected into BEV 2D grid, followed by additional 2D convolutional layers for further feature embedding; finally, a classification branch and localization branch predict the existence of bounding boxes per BEV pixel and bounding box parameters, respectively. cls: classification branch; loc: localization branch.\label{fig:pointpillars}}
  \vspace{-14pt}
\end{figure}

\textit{Metrics.} We use birds-eye view (BEV) and 3D mean average precision (mAP) as metrics. The intersection-over-union (IoU) threshold is 0.7 for vehicles and 0.5 for pedestrians. In addition to overall scores, we provide breakdowns based on the distances between the origin and ground-truth boxes: 0m-30m, 30m-50m, and 50m-infinity (Inf). Following existing works ~\cite{Lang_2019_CVPR,Ngiam2019StarNetTC,Zhou2019EndtoEndMF},  we evaluate our method on bounding boxes that have more than 5 points. 

\textit{Implementation details.} As shown in Figure~\ref{fig:pointpillars}, the object detection model is a variant of PointPillars~\cite{Lang_2019_CVPR}: first, points in 3D are projected to BEV, with a lightweight PointNet~\cite{qi2017pointnet} to aggregate features in each BEV pixel; then, three convolutional blocks are employed to further embed the BEV features; finally, we use two branches to predict bounding box existence and bounding box parameters, respectively. 
The frame rate at test time is 24 FPS on par with PointPillars~\cite{Lang_2019_CVPR}.

\textit{Optimization.} We use Adam~\cite{Kingma2014AdamAM} to train both the multi-frame and the single-frame model. For both models, the initial learning rate is $3\times 10^{-4}$. We then linearly increase it to $3\times10^{-3}$ in the first 5 epochs. Finally it is decreased to $3\times10^{-6}$ using cosine scheduling~\cite{Loshchilov2017SGDRSG}. The model is trained for 75 epochs at each stage. The weight $\lambda$ of feature consistency is 0.1 for vehicles and 0.01 for pedestrians. We use 5 frames for the multi-frame model. Our batch size is 128. We train our models on TPUv3. Training the multi-frame model is about 2 times slower than training the single-frame model.

\subsection{Results on the Waymo Open Dataset}
\label{sec:result}
In Table~\ref{table:car} and Table~\ref{table:ped}, we mainly compare the model trained with our two-stage strategy to a single-frame model trained from scratch. For the sake of completeness, we also provide results of StarNet~\cite{Ngiam2019StarNetTC}, LaserNet~\cite{Meyer2019LaserNetAE}, PointPillars~\cite{Lang_2019_CVPR}, and MVF~\cite{Zhou2019EndtoEndMF}; compared to them, even the single-frame model achieves better performance. The model trained on 5 frames significantly outperforms others, including the single-frame model, but this is an unrealistic setup for testing---points from multiple frames are aggregated by using ground-truth tracking information. The single-frame model with distillation improves over its vanilla counterpart in all metrics, which indicates the model indeed learns additional information with the supervision from a multi-frame model.

\begin{table}[t] 
\begin{center}
\resizebox{\linewidth}{!}{
\begin{tabular}{lcccccccc}
\toprule
\multirow{2}{*}{Method} & \multicolumn{4}{c}{BEV mAP (IoU=0.7)}  & \multicolumn{4}{c}{3D mAP (IoU=0.7)} \\
\cmidrule(lr){2-5}\cmidrule(lr){6-9}
& Overall  &  0 - 30m  & 30 - 50m & 50m - Inf & Overall & 0 - 30m & 30 - 50m & 50m - Inf \\
\midrule
StarNet~\cite{Ngiam2019StarNetTC} & - & - & - & - & 53.7 & - & - & - \\
\cmidrule(lr){2-5}\cmidrule(lr){6-9}
LaserNet$\dag$~\cite{Meyer2019LaserNetAE} & 71.57 & 92.94 & 74.92 & 48.87 & 55.1 & 84.9 & 53.11 & 23.92 \\
\cmidrule(lr){2-5}\cmidrule(lr){6-9}
PointPillars$\ast$~\cite{Lang_2019_CVPR} & 75.57 & 92.1 & 74.06 & 55.47 & 56.62 & 81.01 & 51.75 & 27.94 \\
\cmidrule(lr){2-5}\cmidrule(lr){6-9}
MVF~\cite{Zhou2019EndtoEndMF} & 80.4 & 93.59 & 79.21 & 63.09 & 62.93 & 86.3 & 60.2 & 36.02 \\
\midrule
Baseline (1 frame) & 83.16 & 94.2 & 81.22 & 64.43 &  63.09 & 84.73 & 58.55 & 33.55 \\
\cmidrule(lr){2-5}\cmidrule(lr){6-9}
+ Distillation (1 frame) & \textbf{84.79} & \textbf{94.84} & \textbf{82.69} & \textbf{68.15} & \textbf{66.6} & \textbf{87.56} & \textbf{60.97} & \textbf{38.94} \\
\cmidrule(lr){2-5}\cmidrule(lr){6-9}
Improvements  & \textbf{+1.63} & \textbf{+0.64} & \textbf{+1.47} & \textbf{+3.72} & \textbf{+3.51} & \textbf{+2.83} & \textbf{+2.42} & \textbf{+5.39} \\
\midrule
Oracle (5 frames) $\ddag$  & 86.31 & 95.27 & 84.46 & 70.7 & 68.79 & 87.98 & 65.09 & 41.61 \\
\bottomrule
\end{tabular}
}
\end{center}
\caption{Results on vehicle\label{table:car}. $\dag$: our implementation. $\ast$: re-implemented by~\cite{Zhou2019EndtoEndMF}. $\ddag$: the model trained and tested on dense point clouds as in Figure~\ref{fig:pointclouds} (c). $\ddag$ is not a realistic setup, but we provide results to serve as an oracle.}
\vspace{-20pt}
\end{table}

\begin{table}[t] 
\begin{center}
\resizebox{\linewidth}{!}{
\begin{tabular}{lcccccccc}
\toprule
\multirow{2}{*}{Method} & \multicolumn{4}{c}{BEV mAP (IoU=0.5)}  & \multicolumn{4}{c}{3D mAP (IoU=0.5)} \\ 
\cmidrule(lr){2-5}\cmidrule(lr){6-9}
& Overall  &  0 - 30m  & 30 - 50m & 50m - Inf & Overall & 0 - 30m & 30 - 50m & 50m - Inf \\
\midrule
StarNet~\cite{Ngiam2019StarNetTC} & - & - & - & - & 66.8 & - & - & - \\
\cmidrule(lr){2-5}\cmidrule(lr){6-9}
LaserNet$\dag$~\cite{Meyer2019LaserNetAE}  & 70.01 & 78.24 & 69.47 & 52.68 & 63.4 & 73.47 & 61.55 & 42.69 \\
\cmidrule(lr){2-5}\cmidrule(lr){6-9}
PointPillars$\ast$~\cite{Lang_2019_CVPR} & 68.7 & 75.0 & 66.6 & 58.7 & 60.0 & 68.9 & 57.6 & 46.0 \\
\cmidrule(lr){2-5}\cmidrule(lr){6-9}
MVF~\cite{Zhou2019EndtoEndMF} & 74.38 & 80.01 & 72.98 & 62.51 & 65.33 & 72.51 & 63.35 & 50.62 \\
\midrule
Baseline (1 frame) & 74.48 & 81.72 & 72.9 & 58.77 & 68.32 & 78.12 & 64.85 & 50.21\\
\cmidrule(lr){2-5}\cmidrule(lr){6-9}
+ Distillation (1 frame) & \textbf{76.3} & \textbf{83.03} & \textbf{74.49} & \textbf{62.18} & \textbf{69.9} & \textbf{78.83} & \textbf{66.96} & \textbf{52.73} \\
\cmidrule(lr){2-5}\cmidrule(lr){6-9}
Improvements & \textbf{+1.82} & \textbf{+1.31} & \textbf{+1.59} & \textbf{+3.41} & \textbf{+1.58} & \textbf{+0.71} & \textbf{+2.11} & \textbf{+2.52} \\
\midrule
Oracle (5 frames) $\ddag$ & 80.54 & 86.57 & 78.13 & 69.08 & 74.68 & 82.31 & 71.45 & 61.85 \\
\bottomrule
\end{tabular}
}
\end{center}
\caption{Results on pedestrian\label{table:ped}. $\dag$: our implementation. $\ast$: re-implemented by~\cite{Zhou2019EndtoEndMF}.  $\ddag$: the model trained and tested on dense point clouds as in Figure~\ref{fig:pointclouds} (c). $\ddag$ is not a realistic setup, but we provide results to serve as an oracle.}
\end{table}
{\small
    \bibliographystyle{splncs04}
    \bibliography{distillation}
}
\end{document}